# An Efficient Compiler for Weighted Rewrite Rules

**Mehryar Mohri**
AT&T Research
600 Mountain Avenue
Murray Hill, 07974 NJ
mohri@research.att.com

**Richard Sproat**
Bell Laboratories
700 Mountain Avenue
Murray Hill, 07974 NJ
rws@bell-labs.com

## Abstract

Context-dependent rewrite rules are used in many areas of natural language and speech processing. Work in computational phonology has demonstrated that, given certain conditions, such rewrite rules can be represented as finite-state transducers (FSTs). We describe a new algorithm for compiling rewrite rules into FSTs. We show the algorithm to be simpler and more efficient than existing algorithms. Further, many of our applications demand the ability to compile *weighted* rules into *weighted* FSTs, transducers generalized by providing transitions with weights. We have extended the algorithm to allow for this.

## 1. Motivation

Rewrite rules are used in many areas of natural language and speech processing, including syntax, morphology, and phonology[1]. In interesting applications, the number of rules can be very large. It is then crucial to give a representation of these rules that leads to efficient programs.

Finite-state transducers provide just such a compact representation (Mohri, 1994). They are used in various areas of natural language and speech processing because their increased computational power enables one to build very large machines to model interestingly complex linguistic phenomena. They also allow algebraic operations such as union, composition, and projection which are very useful in practice (Berstel, 1979; Eilenberg, 1974 1976). And, as originally shown by Johnson (1972), rewrite rules can be modeled as finite-state transducers, under the condition that no rule be allowed to apply any more than a finite number of times to its own output.

Kaplan and Kay (1994), or equivalently Karttunen (1995), provide an algorithm for compiling rewrite rules into finite-state transducers, under the condition that they do not rewrite their noncontextual part[2]. We here present a new algorithm for compiling such rewrite rules which is both simpler to understand and implement, and computationally more efficient. Clarity is important since, as pointed out by Kaplan and Kay (1994), the representation of rewrite rules by finite-state transducers involves many subtleties. Time and space efficiency of the compilation are also crucial. Using naive algorithms can be very time consuming and lead to very large machines (Liberman, 1994).

In some applications such as those related to speech processing, one needs to use *weighted* rewrite rules, namely rewrite rules to which weights are associated. These weights are then used at the final stage of applications to output the most probable analysis. Weighted rewrite rules can be compiled into *weighted* finite-state transducers, namely transducers generalized by providing transitions with a weighted output, under the same context condition. These transducers are very useful in speech processing (Pereira et al., 1994). We briefly describe how we have augmented our algorithm to handle the compilation of weighted rules into weighted finite-state transducers.

In order to set the stage for our own contribution, we start by reviewing salient aspects of the Kaplan and Kay algorithm.

[1] Parallel rewrite rules also have interesting applications in biology. In addition to their formal language theory interest, systems such as those of Aristid Lindenmayer provide rich mathematical models for biological development (Rozenberg and Salomaa, 1980).

[2] The general question of the decidability of the halting problem even for one-rule semi-Thue systems is still open. Robert McNaughton (1994) has recently made a positive conjecture about the class of the rules without *self overlap*.

$$
\begin{aligned}
& \mathit{Prologue} \circ & & & (1) \\
Id(\mathit{Obligatory}(\phi, <_i, >)) &= Id(\overline{\Sigma^*_{><\ 0} <_i \phi^0_{><} > \Sigma^*_{><\ 0}}) \circ \\
Id(\mathit{Rightcontext}(\rho, <, >)) &= Id((\overline{\Sigma^*_{>0} > \overline{\rho_{>0}\Sigma^*_{>0} - > \Sigma^*_{>0}}} \cap \overline{\overline{\Sigma^*_{>0} >}\ \rho_{>0}\Sigma^*_{>0} - > \Sigma^*_{>0}})_<) \circ \\
\mathit{Replace} &= [Id(\Sigma^*_{<_i >_i\ 0}) Opt(Id(<_a)\phi^0_{<_c >_c} \times \psi^0_{<_c >_c} Id(>_a))]^* \circ \\
Id(\mathit{Leftcontext}(\lambda, <, >)) &= Id((\overline{\overline{\Sigma^*_{<0}\lambda_{<0} - \Sigma^*_{<0} <} < \Sigma^*_{<0}} \cap \overline{\Sigma^*_{<0}\lambda_{<0} - \Sigma^*_{<0} < \overline{< \Sigma^*_{<0}}})_>) \circ \\
& \mathit{Prologue}^{-1}
\end{aligned}
$$

Figure 1: Compilation of obligatory left-to-right rules, using the KK algorithm.

## 2. The KK Algorithm

The rewrite rules we consider here have the following general form:

$$\phi \rightarrow \psi / \lambda ___ \rho \tag{2}$$

Such rules can be interpreted in the following way: $\phi$ is to be replaced by $\psi$ whenever it is preceded by $\lambda$ and followed by $\rho$. Thus, $\lambda$ and $\rho$ represent the left and right contexts of application of the rules. In general, $\phi$, $\psi$, $\lambda$ and $\rho$ are all regular expressions over the alphabet of the rules. Several types of rules can be considered depending on their being obligatory or optional, and on their direction of application, from left to right, right to left or simultaneous application.

Consider an obligatory rewrite rule of the form $\phi \rightarrow \psi / \lambda ___ \rho$, which we will assume applies left to right across the input string. Compilation of this rule in the algorithm of Kaplan and Kay (1994) (KK for short) involves composing together six transducers, see Figure 1.

We use the notations of KK. In particular, $\Sigma$ denotes the alphabet, $<$ denotes the set of context labeled brackets $\{<_a, <_i, <_c\}$, $>$ the set $\{>_a, >_i, >_c\}$, and 0 an additional character representing deleted material. Subscript symbols of an expression are symbols which are allowed to freely appear anywhere in the strings represented by that expression. Given a regular expression $r$, $Id(r)$ is the identity transducer obtained from an automaton $A$ representing $r$ by adding output labels to $A$ identical to its input labels.

The first transducer, *Prologue*, freely introduces labeled brackets from the set $\{<_a, <_i, <_c, >_a, >_i, >_c\}$ which are used by left and right context transducers. The last transducer, $\mathit{Prologue}^{-1}$, erases all such brackets.

In such a short space, we can of course not hope to do justice to the KK algorithm, and the reader who is not familiar with it is urged to consult their paper. However, one point that we do need to stress is the following: while the construction of *Prologue*, $\mathit{Prologue}^{-1}$ and *Replace* is fairly direct, construction of the other transducers is more complex, with each being derived via the application of several levels of regular operations from the original expressions in the rules. This clearly appears from the explicit expressions we have indicated for the transducers. The construction of the three other transducers involves many operations including: two intersections of automata, two distinct subtractions, and nine complementations. Each subtraction involves an intersection and a complementation algorithm[3]. So, in the whole, four intersections and eleven complementations need to be performed.

Intersection and complementation are classical automata algorithms (Aho et al., 1974; Aho et al., 1986). The complexity of intersection is quadratic. But the classical complementation algorithm requires the input automaton to be deterministic. Thus, each of these 11 operations requires first the determinization of the input. Such operations can be very costly in the case of the automata involved in the KK algorithm[4].

In the following section we briefly describe a new algorithm for compiling rewrite rules. For reasons of space, we concentrate here on the compilation of left-to-right obligatory rewrite rules. However, our methods extend straightforwardly to other modes of application (optional, right-to-left, simultaneous, batch), or kinds of rules (two-level rules) discussed by Kaplan and Kay (1994).

[3] A subtraction can of course also be performed *directly* by combining the two steps of intersection and complementation, but the corresponding algorithm has exactly the same cost as the total cost of the two operations performed consecutively.

[4] One could hope to find a more efficient way of determining the complement of an automaton that would not require determinization. However, this problem is PSPACE-complete. Indeed, the *regular expression non-universality* problem is a subproblem of complementation known to be PSPACE-complete (Garey and Johnson, 1979, page 174), (Stockmeyer and Meyer, 1973). This problem also known as *the emptiness of complement problem* has been extensively studied (Aho et al., 1974, page 410-419).

# 3. New Algorithm

## 3.1. Overview

In contrast to the KK algorithm which introduces brackets everywhere only to restrict their occurrence subsequently, our algorithm introduces context symbols just when and where they are needed. Furthermore, the number of intermediate transducers necessary in the construction of the rules is smaller than in the KK algorithm, and each of the transducers can be constructed more directly and efficiently from the primitive expressions of the rule, $\phi$, $\psi$, $\lambda$, $\rho$.

A transducer corresponding to the left-to-right obligatory rule $\phi \rightarrow \psi/\lambda \underline{\quad} \rho$ can be obtained by composition of five transducers:

$$r \circ f \circ replace \circ l_1 \circ l_2 \qquad (3)$$

1. The transducer $r$ introduces in a string a marker $>$ before every instance of $\rho$. For reasons that will become clear we will notate this as $\Sigma^* \rho \rightarrow \Sigma^* > \rho$.
2. The transducer $f$ introduces markers $<_1$ and $<_2$ before each instance of $\phi$ that is followed by $>$: $(\Sigma \cup \{>\})^* \phi > \rightarrow (\Sigma \cup \{>\})^* \{<_1, <_2\} \phi >$. In other words, this transducer marks just those $\phi$ that occur before $\rho$.
3. The replacement transducer *replace* replaces $\phi$ with $\psi$ in the context $<_1 \phi >$, simultaneously deleting $>$ in all positions (Figure 2). Since $>$, $<_1$, and $<_2$ need to be ignored when determining an occurrence of $\phi$, there are loops over the transitions $>: \epsilon, <_1: \epsilon, <_2: \epsilon$ at all states of $\phi$, or equivalently of the states of the cross product transducer $\phi \times \psi$.
4. The transducer $l_1$ admits only those strings in which occurrences of $<_1$ are preceded by $\lambda$ and deletes $<_1$ at such occurrences: $\Sigma^* \lambda <_1 \rightarrow \Sigma^* \lambda$.
5. The transducer $l_2$ admits only those strings in which occurrences of $<_2$ are not preceded by $\overline{\lambda}$ and deletes $<_2$ at such occurrences: $\Sigma^* \overline{\lambda} <_2 \rightarrow \Sigma^* \overline{\lambda}$.

Clearly the composition of these transducers leads to the desired result. The construction of the transducer *replace* is straightforward. In the following, we show that the construction of the other four transducers is also very simple, and that it only requires the determinization of 3 automata and additional work linear (time and space) in the size of the determinized automata.

## 3.2. Markers

### Markers of TYPE 1

Let us start by considering the problem of constructing what we shall call a TYPE 1 transducer, which inserts a marker after all prefixes of a string that match a particular regular expression. Given a regular expression $\beta$ defined on the alphabet $\Sigma$, one can construct, using classical algorithms (Aho et al., 1986), a deterministic automaton $\alpha$ representing $\Sigma^* \beta$. As with the KK algorithm, one can obtain from $\alpha$ a transducer $\chi = Id(\alpha)$ simply by assigning to each transition the same output label as the input label. We can easily transform $\chi$ into a new transducer $\tau$ such that it inserts an arbitrary marker # after each occurrence of a pattern described by $\beta$. To do so, we make final the non-final states of $\chi$ and for any final state $q$ of $\chi$ we create a new state $q'$, a copy of $q$. Thus, $q'$ has the same transitions as $q$, and $q'$ is a final state. We then make $q$ non-final, remove the transitions leaving $q$ and add a transition from $q$ to $q'$ with input label the empty word $\epsilon$, and output #. Figures 3 and 4 illustrate the transformation of $\chi$ into $\tau$.

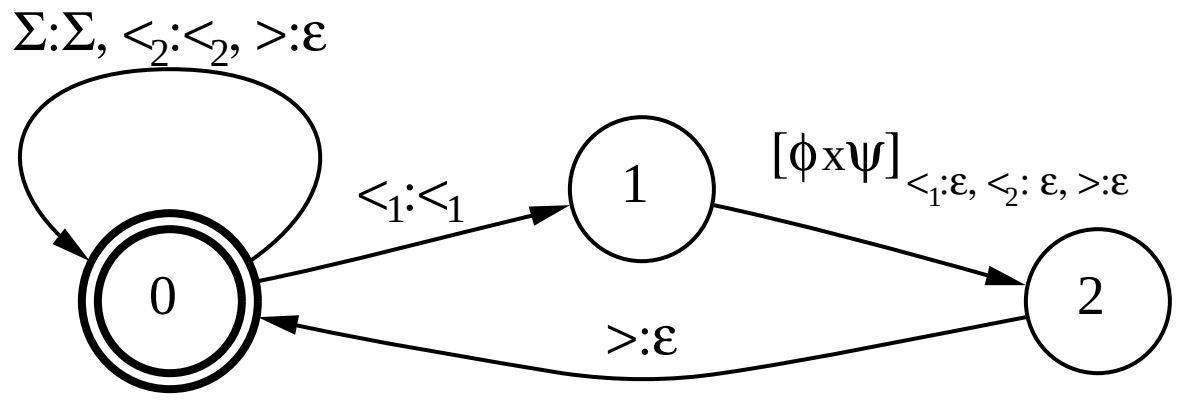

Figure 2: Replacement transducer *replace* in the obligatory left-to-right case.

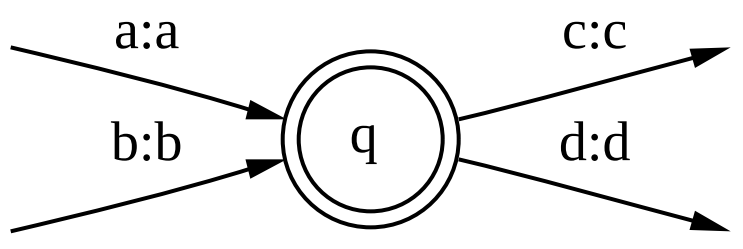

Figure 3: Final state $q$ of $\chi$ with entering and leaving transitions.

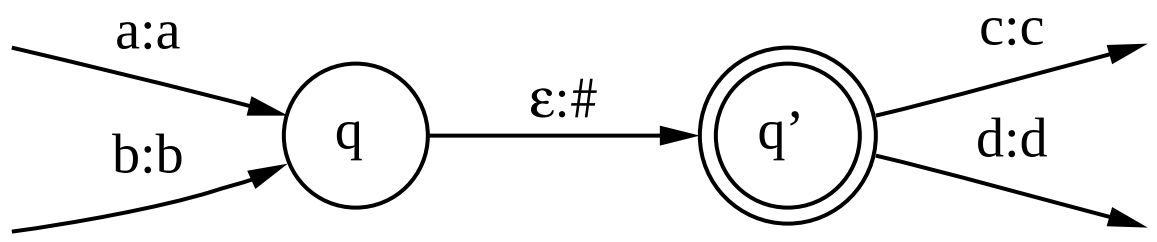

Figure 4: States and transitions of $\tau$ obtained by modifications of those of $\chi$.

**Proposition 1** *Let $\alpha$ be a deterministic automaton representing $\Sigma^*\beta$, then the transducer $\tau$ obtained as described above is a transducer post-marking occurrences of $\beta$ in a string of $\Sigma^*$ by #.*

Proof. The proof is based on the observation that a deterministic automaton representing $\Sigma^*\beta$ is necessarily complete[5]. Notice that non-deterministic automata representing $\Sigma^*\beta$ are not necessarily complete. Let $q$ be a state of $\alpha$ and let $u \in \Sigma^*$ be a string reaching $q$[6]. Let $v$ be a string described by the regular expression $\beta$. Then, for any $a \in \Sigma$, $uav$ is in $\Sigma^*\beta$. Hence, $uav$ is accepted by the automaton $\alpha$, and, since $\alpha$ is deterministic, there exists a transition labeled with $a$ leaving $q$. Thus, one can read any string $u \in \Sigma^*$ using the automaton $\alpha$. Since by definition of $\alpha$, the state reached when reading a prefix $u'$ of $u$ is final iff $u' \in \Sigma^*\beta$, by construction, the transducer $\tau$ inserts the symbol # after the prefix $u'$ iff $u'$ ends with a pattern of $\beta$. This ends the proof of the proposition. □

**Markers of TYPE 2**

In some cases, one wishes to check that any occurrence of # in a string $s$ is preceded (or followed) by an occurrence of a pattern of $\beta$. We shall say that the corresponding transducers are of TYPE 2. They play the role of a filter. Here again, they can be defined from a deterministic automaton representing $\Sigma^*\beta$. Figure 5 illustrates the modifications to make from the automaton of figure 3. The symbols # should only appear at final states and must be erased. The loop $\# : \epsilon$ added at final states of $Id(\alpha)$ is enough for that purpose. All states of the transducer are then made final since any string conforming to this restriction is acceptable: cf. the transducer $l_1$ for $\lambda$ above.

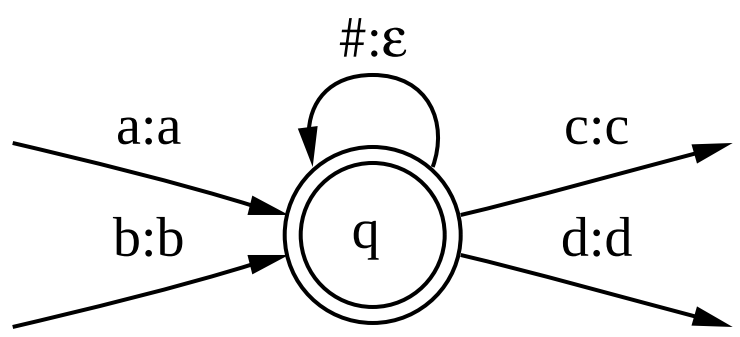

Figure 5: Filter transducer, TYPE 2.

[5]An automaton $A$ is complete iff at any state $q$ and for any element $a$ of the alphabet $\Sigma$ there exists at least one transition leaving $q$ labeled with $a$. In the case of deterministic automata, the transition is unique.

[6]We assume all states of $\alpha$ accessible. This is true if $\alpha$ is obtained by determinization.

**Markers of TYPE 3**

In other cases, one wishes to check the reverse constraint, that is that occurrences of # in the string $s$ are not preceded (or followed) by any occurrence of a pattern of $\beta$. The transformation then simply consists of adding a loop at each non-final state of $Id(\alpha)$, and of making all states final. Thus, a state such as that of figure 6 is transformed into that of figure 5. We shall say that the corresponding transducer is of TYPE 3: cf. the transducer $l_2$ for $\overline{\lambda}$.

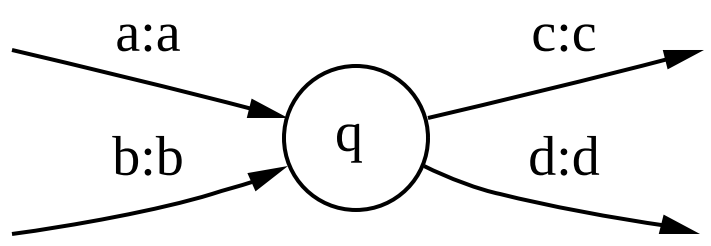

Figure 6: Non-final state $q$ of $\alpha$.

The construction of these transducers (TYPE 1-3) can be generalized in various ways. In particular:

- One can add several alternative markers $\{\#_1, \cdots, \#_k\}$ after each occurrence of a pattern of $\beta$ in a string. The result is then an automaton with transitions labeled with, for instance, $\#_1, \cdots, \#_k$ after each pattern of $\beta$: cf. transducer $f$ for $\phi$ above.
- Instead of inserting a symbol, one can delete a symbol which would be necessarily present after each occurrence of a pattern of $\beta$.

For any regular expression $\alpha$, define $Marker(\alpha, type, deletions, insertions)$ as the transducer of type *type* constructed as previously described from a deterministic automaton representing $\alpha$, *insertions* and *deletions* being, respectively, the set of insertions and deletions the transducer makes.

**Proposition 2** *For any regular expression $\alpha$, $Marker(\alpha, type, deletions, insertions)$ can be constructed from a deterministic automaton representing $\alpha$ in linear time and space with respect to the size of this automaton.*

Proof. We proved in the previous proposition that the modifications do indeed lead to the desired transducer for TYPE 1. The proof for other cases is similar. That the construction is linear in space is clear since at most one additional transition and state is created for final or non-final states[7]. The overall time complexity of the construction is linear, since the construction of $Id(\alpha)$ is linear in the

[7]For TYPE 2 and TYPE 3, no state is added but only a transition per final or non-final state.

$$r = [reverse(Marker(\Sigma^* reverse(\rho), 1, \{>\}, \emptyset))] \quad (4)$$
$$f = [reverse(Marker((\Sigma \cup \{>\})^* reverse(\phi_> >), 1, \{<_1, <_2\}, \emptyset))] \quad (5)$$
$$l_1 = [Marker(\Sigma^* \lambda, 2, \emptyset, \{<_1\})]_{<_2:<_2} \quad (6)$$
$$l_2 = [Marker(\Sigma^* \lambda, 3, \emptyset, \{<_2\})] \quad (7)$$

Figure 7: Expressions of the $r$, $f$, $l_1$, and $l_2$ using $Marker$.

number of transitions of $\alpha$ and that other modifications consisting of adding new states and transitions and making states final or not are also linear. □

We just showed that $Marker(\alpha, type, deletions, insertions)$ can be constructed in a very efficient way. Figure 7 gives the expressions of the four transducers $r$, $f$, $l_1$, and $l_2$ using $Marker$.

Thus, these transducers can be constructed very efficiently from deterministic automata representing[8] $\Sigma^* reverse(\rho)$, $(\Sigma \cup \{>\})^* reverse(\phi_> >)$, and $\Sigma^* \lambda$. The construction of $r$ and $f$ requires two reverse operations. This is because these two transducers insert material *before* $\rho$ or $\phi$.

## 4. Extension to Weighted Rules

In many applications, in particular in areas related to speech, one wishes not only to give all possible analyses of some input, but also to give some measure of how likely each of the analyses is. One can then generalize replacements by considering *extended* regular expressions, namely, using the terminology of formal language theory, *rational power series* (Berstel and Reutenauer, 1988; Salomaa and Soittola, 1978).

The rational power series we consider here are functions mapping $\Sigma^*$ to $\mathcal{R}_+ \cup \{\infty\}$ which can be described by regular expressions over the alphabet $(\mathcal{R}_+ \cup \{\infty\}) \times \Sigma$. $S = (4a)(2b)^*(3b)$ is an example of rational power series. It defines a function in the following way: it associates a non-null number only with the strings recognized by the regular expression $ab^*b$. This number is obtained by adding the coefficients involved in the recognition of the string. The value associated with $abbb$, for instance, is $(S, abbb) = 4 + 2 + 2 + 3 = 11$.

In general, such extended regular expressions can be redundant. Some strings can be matched in different ways with distinct coefficients. The value associated with those strings is then the minimum of all possible results. $S' = (2a)(3b)(4b) + (5a)(3b^*)$ matches $abb$ with the different weights $2 + 3 + 4 = 9$ and $5 + 3 + 3 = 11$. The minimum of the two is the value associated with $abb$: $(S', abb) = 9$. Non-negative numbers in the definition of these power series are often interpreted as the negative logarithm of probabilities. This explains our choice of the operations: addition of the weights along the string recognition and $min$, since we are only interested in that result which has the highest probability[9].

Rewrite rules can be generalized by letting $\psi$ be a rational power series. The result of the application of a generalized rule to a string is then a set of *weighted strings* which can be represented by a *weighted automaton*. Consider for instance the following rule, which states that an abstract nasal, denoted $N$, is rewritten as $m$ in the context of a following labial:

$$N \rightarrow m / \underline{\quad}[+labial] \quad (8)$$

Now suppose that this is only *probabilistically* true, and that while ninety percent of the time $N$ does indeed become $m$ in this environment, about ten percent of the time in real speech it becomes $n$. Converting from probabilities to weights, one would say that $N$ becomes $m$ with weight $\alpha = -\log(0.9)$, and $n$ with weight $\beta = -\log(0.1)$, in the stated environment. One could represent this by the following rule:

$$N \rightarrow \alpha m + \beta n / \underline{\quad}[+labial] \quad (9)$$

We define *Weighted finite-state transducers* as transducers such that in addition to input and output labels, each transition is labeled with a weight.

The result of the application of a weighted transducer to a string, or more generally to an automaton is a weighted automaton. The corresponding operation is similar to the unweighted case. However, the weight of the transducer and those of the string or automaton need to be combined too, here added, during composition (Pereira et al., 1994).

[8] As in the KK algorithm we denote by $\phi_>$ the set of the strings described by $\phi$ containing possibly occurrences of $>$ at any position. In the same way, subscripts such as $>:>$ for a transducer $\tau$ indicate that loops by $>:>$ are added at all states of $\tau$. We denote by $reverse(\alpha)$ the regular expression describing exactly the reverse strings of $\alpha$ if $\alpha$ is a regular expression, or the reverse transducer of $\alpha$ if $\alpha$ is a transducer.

[9] Using the terminology of the theory of languages, the functions we consider here are power series defined on the *tropical semiring* $(\mathcal{R}_+ \cup \{\infty\}, \min, +, \infty, 0)$ (Kuich and Salomaa, 1986).

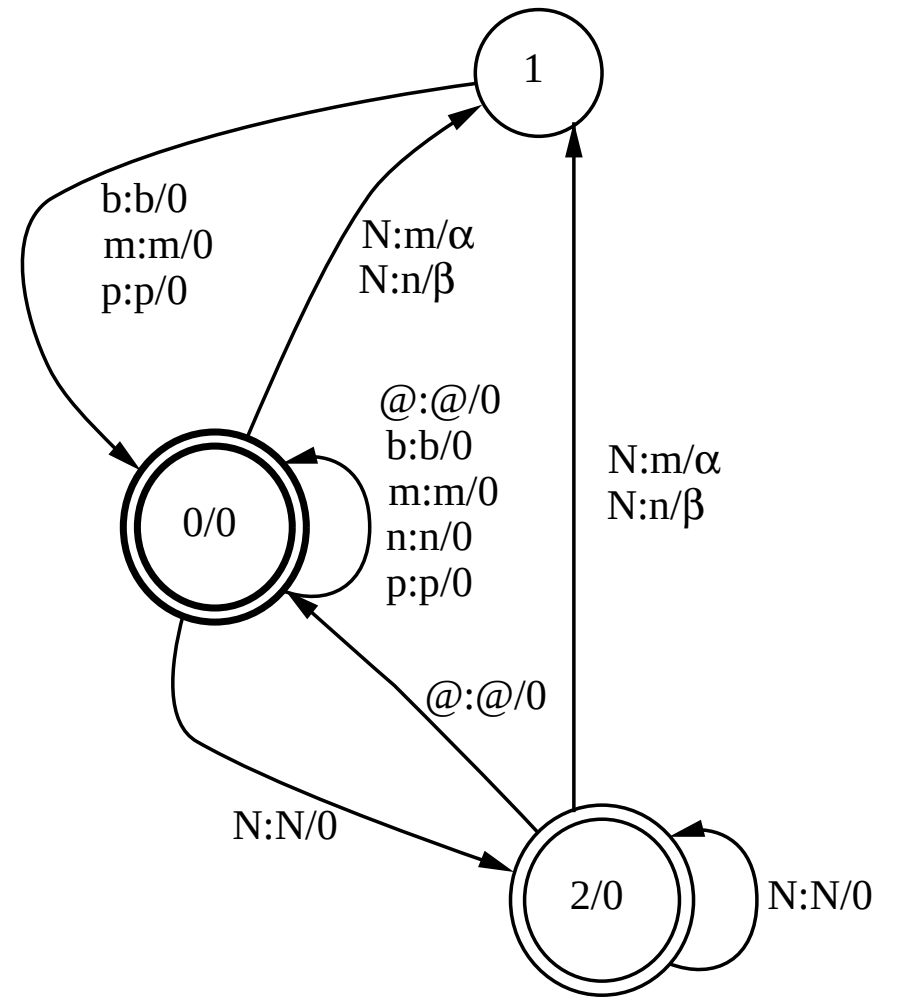

Figure 8: Transducer representing the rule 9.

We have generalized the composition operation to the weighted case by introducing this combination of weights. The algorithm we described in the previous sections can then also be used to compile weighted rewrite rules.

As an example, the obligatory rule 9 can be represented by the weighted transducer of Figure 8 [10]. The following theorem extends to the weighted case the assertion proved by Kaplan and Kay (1994).

**Theorem 1** *A weighted rewrite rule of the type defined above that does not rewrite its non-contextual part can be represented by a weighted finite-state transducer.*

Proof. The construction we described in the previous section also provides a constructive proof of this theorem in the unweighted case. In case $\psi$ is a power series, one simply needs to use in that construction a weighted finite-state transducer representing $\psi$. By definition of composition of weighted transducers, or multiplication of power series, the weights are then used in a way consistent with the definition of the weighted context-dependent rules. □

## 5. Experiments

In order to compare the performance of the algorithm presented here with KK, we timed both algorithms on the compilation of individual rules taken from the following set ($k \in [0, 10]$):

$$a \rightarrow b/\ c^k ____ \qquad (10)$$

$$a \rightarrow b/\ ____ c^k \qquad (11)$$

[10]We here use the symbol @ to denote all letters different from $b$, $m$, $n$, $p$, and $N$.

In other words we tested twenty two rules where the left context or the right context is varied in length from zero to ten occurrences of $c$. For our experiments, we used the alphabet of a realistic application, the text analyzer for the Bell Laboratories German text-to-speech system consisting of 194 labels. All tests were run on a Silicon Graphics IRIS Indigo 4000, 100 MhZ IP20 Processor, 128 Mbytes RAM, running IRIX 5.2. Figure 9 shows the relative performance of the two algorithms for the left context: apparently the performance of both algorithms is *roughly* linear in the length of the left context, but KK has a worse constant, due to the larger number of operations involved. Figure 10 shows the equivalent data for the right context. At first glance the data looks similar to that for the left context, until one notices that in Figure 10 we have plotted the time on a log scale: the KK algorithm is hyperexponential.

What is the reason for this performance degradation in the right context? The culprits turn out to be the two intersectands in the expression of $Rightcontext(\rho, <, >)$ in Figure 1. Consider for example the righthand intersectand, namely $\overline{\overline{\Sigma^*_{>0} >}\ \rho_{>0}\Sigma^*_{>0} - > \Sigma^*_{>0}}$, which is the complement of $\overline{\Sigma^*_{>0} >}\ \rho_{>0}\Sigma^*_{>0} - > \Sigma^*_{>0}$. As previously indicated, the complementation algorithm requires determinization, and the determinization of automata representing expressions of the form $\Sigma^*\alpha$, where $\alpha$ is a regular expression, is often very expensive, specially when the expression $\alpha$ is already complex, as in this case.

Figure 11 plots the behavior of determinization on the expression $\overline{\Sigma^*_{>0} >}\ \rho_{>0}\Sigma^*_{>0} - > \Sigma^*_{>0}$ for each of the rules in the set $a \rightarrow b/____ c^k$, ($k \in [0, 10]$). On the horizontal axis is the number of arcs of the non-deterministic input machine, and on the vertical axis the log of the number of arcs of the deterministic machine, i.e. the machine result of the determinization algorithm without using any minimization. The perfect linearity indicates an exponential time and space behavior, and this in turn explains the observed difference in performance. In contrast, the construction of the right context machine in our algorithm involves only the single determinization of the automaton representing $\Sigma^*\rho$, and thus is much less expensive. The comparison just discussed involves a rather artificial ruleset, but the differences in performance that we have highlighted show up in real applications. Consider two sets of pronunciation rules from the Bell Laboratories German text-to-speech system: the size of the alphabet for this ruleset is 194, as noted above. The first ruleset, consisting of pronunciation rules for the orthographic vowel <ö> contains twelve rules, and the second ruleset, which deals with the orthographic vowel <a> contains twenty five rules. In the actual application of the rule compiler to these rules, one compiles the individual rules in each ruleset one by one, and composes them together in the order written, compacts them after each composition, and derives a single transducer for each set. When done off-line, these operations of composition and compaction dominate the time corresponding to the construction of the transducer for each individual rule. The difference between the two algorithms appears still clearly for these two sets of rules. Table 1 shows for each algorithm the times in seconds for the overall construction, and the number of states and arcs of the output transducers.

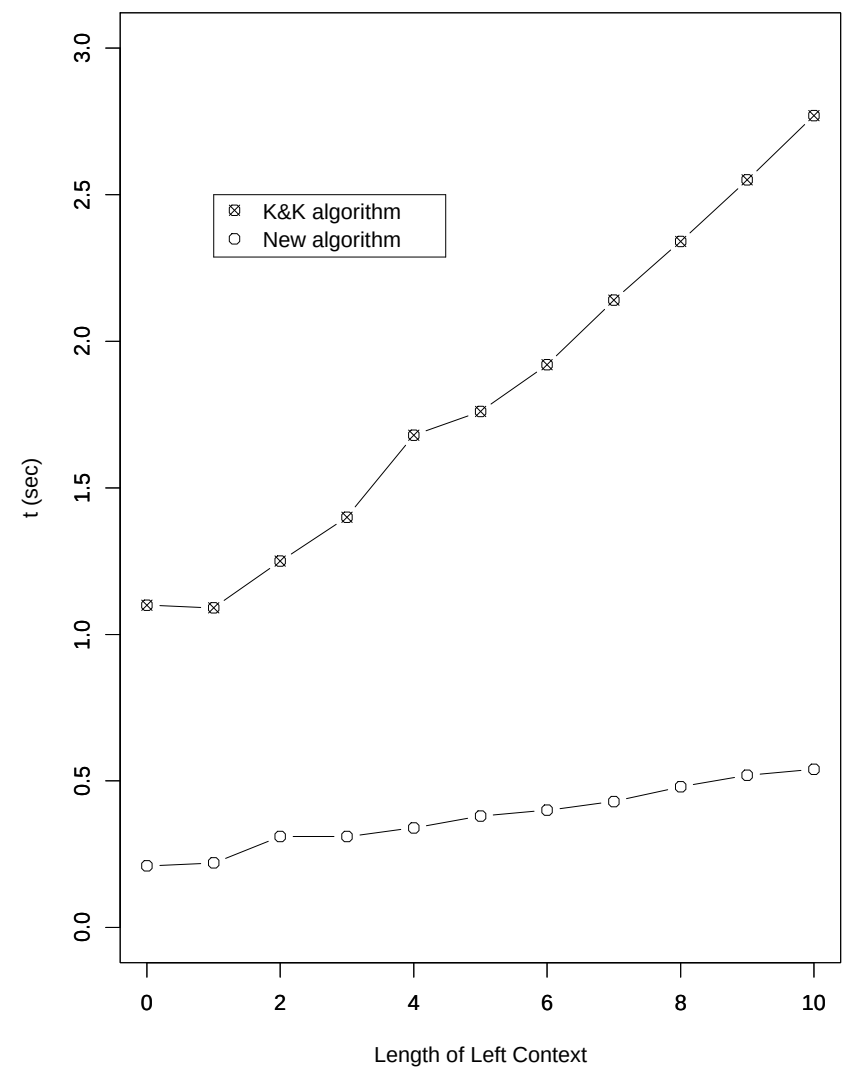

Figure 9: Compilation times for rules of the form $a \rightarrow b/\ c^k___$, ($k \in [0, 10]$).

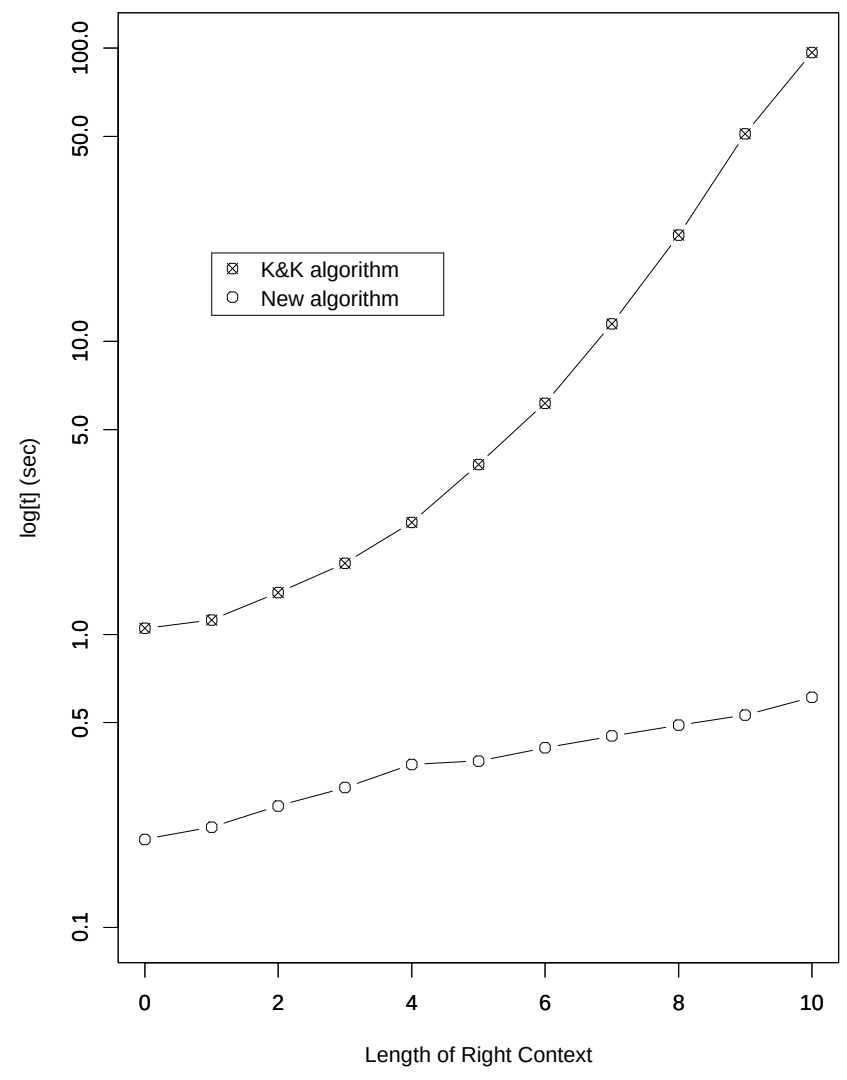

Figure 10: Compilation times for rules of the form $a \rightarrow b/___c^k$, ($k \in [0, 10]$).

Table 1: Comparison in a real example.

| Rules | KK | | | New | | |
|---|---|---|---|---|---|---|
| | time | space | | time | space | |
| | (s) | states | arcs | (s) | states | arcs |
| <ö> | 62 | 412 | 50,475 | 47 | 394 | 47,491 |
| <a> | 284 | 1,939 | 215,721 | 240 | 1,927 | 213,408 |

## 6. Conclusion

We briefly described a new algorithm for compiling context-dependent rewrite rules into finite-state transducers. Several additional methods can be used to make this algorithm even more efficient.

The automata determinizations needed for this algorithm are of a specific type. They represent expressions of the type $\Sigma^*\phi$ where $\phi$ is a regular expression. Given a deterministic automaton representing $\phi$, such determinizations can be performed in a more efficient way using *failure functions* (Mohri, 1995). Moreover, the corresponding determinization is independent of $\Sigma$ which can be very large in some applications. It only depends on the alphabet of the automaton representing $\phi$.

One can devise an *on-the-fly* implementation of the composition algorithm leading to the final transducer representing a rule. Only the necessary part of the intermediate transducers is then expanded for a given input (Pereira et al., 1994).

The resulting transducer representing a rule is often subsequentiable or $p$-subsequentiable. It can then be determinized and minimized (Mohri, 1994). This both makes the use of the transducer time efficient and reduces its size.

We also indicated an extension of the theory of rule-compilation to the case of weighted rules, which compile into weighted finite-state transducers. Many algorithms used in the finite-state theory and in their applications to natural language processing can be extended in the same way.

To date the main serious application of this compiler has been to developing text-analyzers for text-to-speech systems at Bell Laboratories (Sproat, 1996): partial to more-or-less complete analyzers have been built for Spanish, Italian, French, Romanian, German, Russian, Mandarin and Japanese. However, we hope to also be able to use the compiler in serious applications in speech

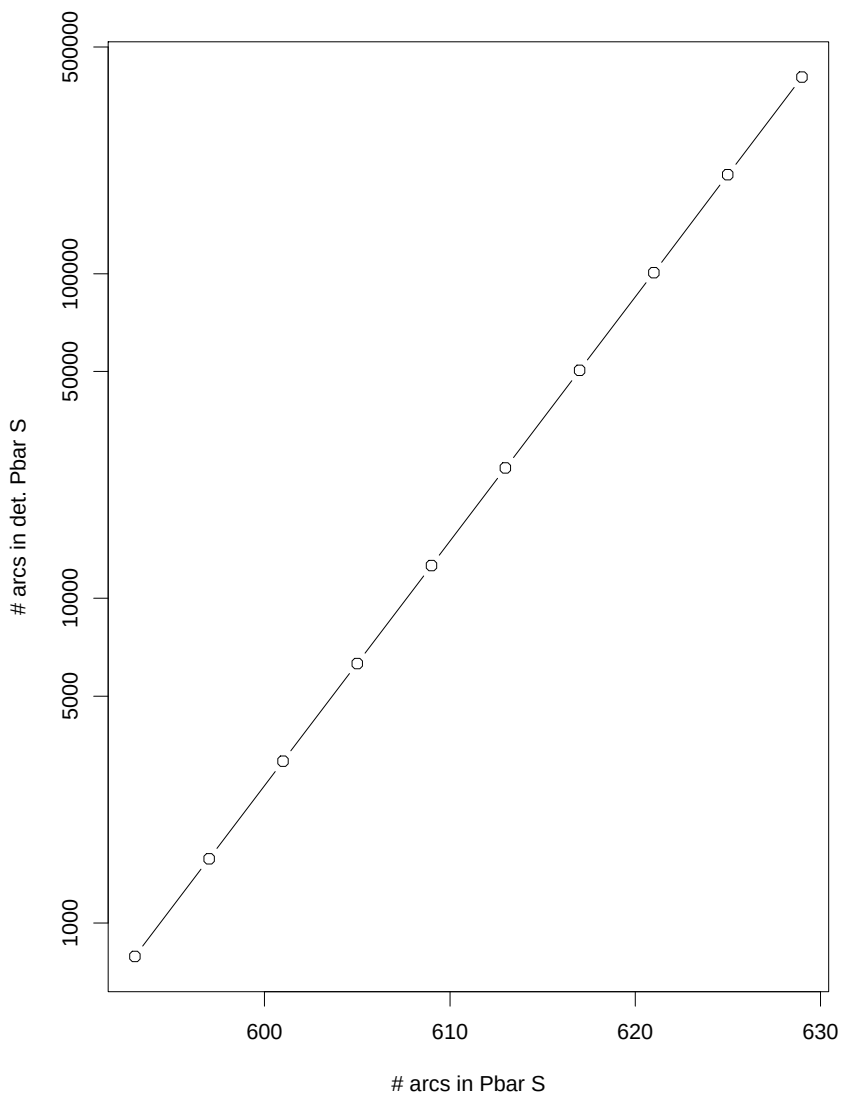

Figure 11: Number of arcs in the non-deterministic automaton $\tau$ representing $\overline{PS} = \overline{\Sigma^*_{>0}} > \rho_{>0}\Sigma^*_{>0} - > \Sigma^*_{>0}$ versus the log of the number of arcs in the automaton obtained by determinization of $\tau$.

recognition in the future.

## Acknowledgements

We wish to thank several colleagues of AT&T/Bell Labs, in particular Fernando Pereira and Michael Riley for stimulating discussions about this work and Bernd Möbius for providing the German pronunciation rules cited herein.